\documentclass[letterpaper, 10 pt, journal, twoside]{ieeetran}

\IEEEoverridecommandlockouts                              

\usepackage{graphics} 
\usepackage{epsfig} 
\usepackage{times} 
\usepackage{amsmath} 
\usepackage{amssymb}  
\usepackage{subfig}

\usepackage[pagebackref=true,breaklinks=true,colorlinks,bookmarks=false]{hyperref}

\newcommand{\tablestyle}[2]{\setlength{\tabcolsep}{#1}\renewcommand{\arraystretch}{#2}\centering\small}
\newcommand{\dt}[1]{\fontsize{7pt}{0.1em}\selectfont (#1)}
\newcommand{\bd}[1]{\textbf{#1}}
\newlength\savewidth\newcommand\shline{\noalign{\global\savewidth\arrayrulewidth
  \global\arrayrulewidth 1pt}\hline\noalign{\global\arrayrulewidth\savewidth}}

\title{\LARGE \bf
Multi-Scale Feature Aggregation by Cross-Scale Pixel-to-Region Relation Operation for Semantic Segmentation
}

\author{Yechao Bai$^{1}$, Ziyuan Huang$^{1}$, \\ 
Lyuyu Shen$^{1}$, Hongliang Guo$^{2}$, Marcelo H. Ang Jr$^{1}$ and Daniela Rus$^{3}$
\thanks{*This work was supported by the National Research Foundation, Prime Minister's Office, Singapore, under its CREATE program, Singapore-MIT Alliance for Research and Technology (SMART) Future Urban Mobility (FM) IRG. (\textit{Yechao Bai and Ziyuan Huang are co-first authors.}) (\textit{Corresponding authors: Yechao Bai.})}

\thanks{$^{1}$Yechao Bai, Ziyuan Huang, Lyuyu Shen and Marcelo H. Ang Jr are with the Department of Mechanical Engineering, National University of Singapore, Singapore.
        {\tt\footnotesize \{yechao.bai, ziyuan.huang, e0444150\}@u.nus.edu, mpeangh@nus.edu.sg}
        }%
\thanks{$^{2}$Hongliang Guo is with the Singapore-MIT Alliance for Research and Technology (SMART) Centre, Singapore.
        {\tt\footnotesize hongliang@smart.mit.edu}}%
\thanks{$^{3}$Daniela Rus is with the Massachusetts Institute of Technology, Cambridge, MA, USA.
        {\tt\footnotesize rus@csail.mit.edu}}%
}

\begin{document}

\IEEEoverridecommandlockouts
\IEEEpubid{\makebox[\columnwidth]{978-1-5386-5541-2/18/\$31.00~\copyright2018 IEEE \hfill} \hspace{\columnsep}\makebox[\columnwidth]{ }}
\maketitle
\IEEEpubidadjcol

\begin{abstract}
Exploiting multi-scale features has shown great potential in tackling semantic segmentation problems.
The aggregation is commonly done with sum or concatenation (concat) followed by convolutional (conv) layers. However, it fully passes down the high-level context to the following hierarchy without considering their interrelation.
In this work, we aim to enable the low-level feature to aggregate the complementary context from adjacent high-level feature maps by a cross-scale pixel-to-region relation operation. We leverage cross-scale context propagation to make the long-range dependency capturable even by the high-resolution low-level features. 
To this end, we employ an efficient feature pyramid network to obtain multi-scale features. We propose a Relational Semantics Extractor (RSE) and Relational Semantics Propagator (RSP) for context extraction and propagation respectively. Then we stack several RSP into an RSP head to achieve the progressive top-down distribution of the context.
Experiment results on two challenging datasets Cityscapes and COCO demonstrate that the RSP head performs competitively on both semantic segmentation and panoptic segmentation with high efficiency. It outperforms DeeplabV3 \cite{deeplabv3} by 0.7\% with 75\% fewer FLOPs (multiply-adds) in the semantic segmentation task.
\end{abstract}


\bstctlcite{IEEEexample:BSTcontrol}

\section{INTRODUCTION}
\begin{figure}[t]
\begin{center}
  \includegraphics[width=1.0 \linewidth]{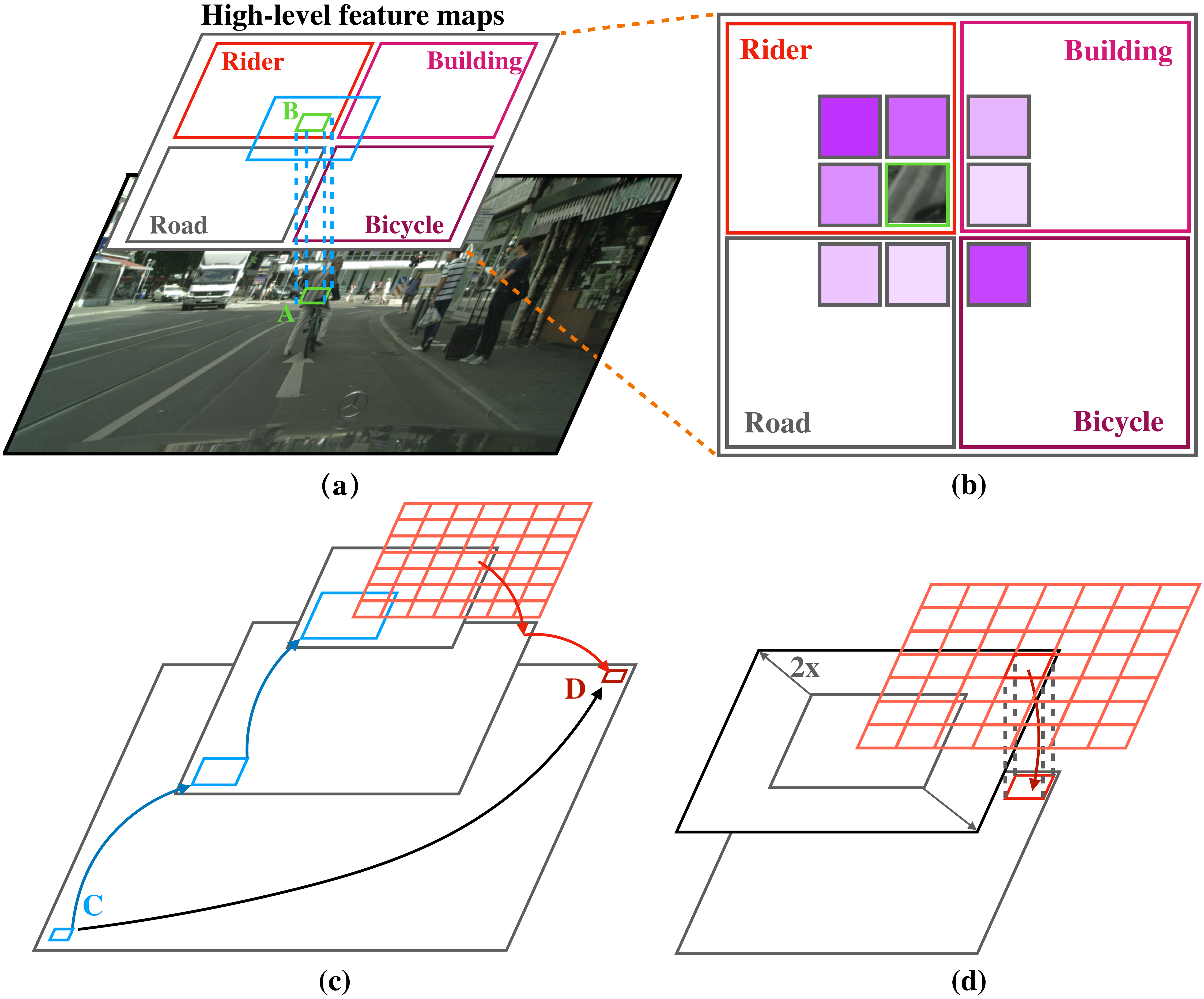}
\end{center}\vspace{-4mm}
  \caption{(a) Cross-scale pixel-to-region relation. This demonstrates that not all context from high-level feature maps is beneﬁcial to the classification of the small portion of a rider in A. (b) The proposed relation operation emphasizes the related context (deeper purple) and suppresses the unrelated context (lighter purple) from the corresponding region of the high-level feature map. (c) The blue arrows represent the context extraction. The pink grids indicate the region that the feature on the adjacent low-level feature maps search and aggregate complementary context. The red arrows represent the propagation of the context. The black arrow implies that feature D in essence captures long-range dependencies from feature C. (d) The high-level feature map is firstly upsampled to the same spatial dimension as the adjacent low-level feature map. The region on the upsampled high-level feature map is centered at the feature at the same spatial position as the low-level feature.}
\label{fig:story}
\end{figure}
Semantic segmentation is a fundamental task in computer vision that has various important applications in self-driving car, robotics, etc. Great advancement has been achieved since the advent of deep neural networks. 
Lots of works have shown that effective integration of contextual information plays a central role in pushing forward the segmentation performance \cite{parsenet, pspnet, deeplabv3+, acfnet, ocr, hierarchy, shape-variant-context, ccnet}.
Contextual information implies the relational connection between an object and a region which facilitates the classification of the object.

As the output of the layers of the CNN backbone encodes different scales and levels of contextual information which combine to form a feature pyramid, it emerges as a natural choice to leverage this multi-scale feature pyramid to achieve a high quality yet efficient context fusion.
The multi-scale feature aggregation is commonly done with sum or concat followed by conv layers with a pixel-to-pixel correspondence.
However, it fully passes down the high-level context to the following hierarchy without considering their interrelation.
For example in Fig. \ref{fig:story} (a) and (b), not all context information in the predefined vicinity of the corresponding high-level feature B is beneficial to the classification of the low-level feature A (a portion of a rider). Ideally, feature A should discriminately aggregate feature that contains the high-level context, emphasize the semantically related and spatially close features from Rider and Bicycle, and suppress the others.

\begin{figure}[t]
\begin{center}
  \includegraphics[width=1\linewidth]{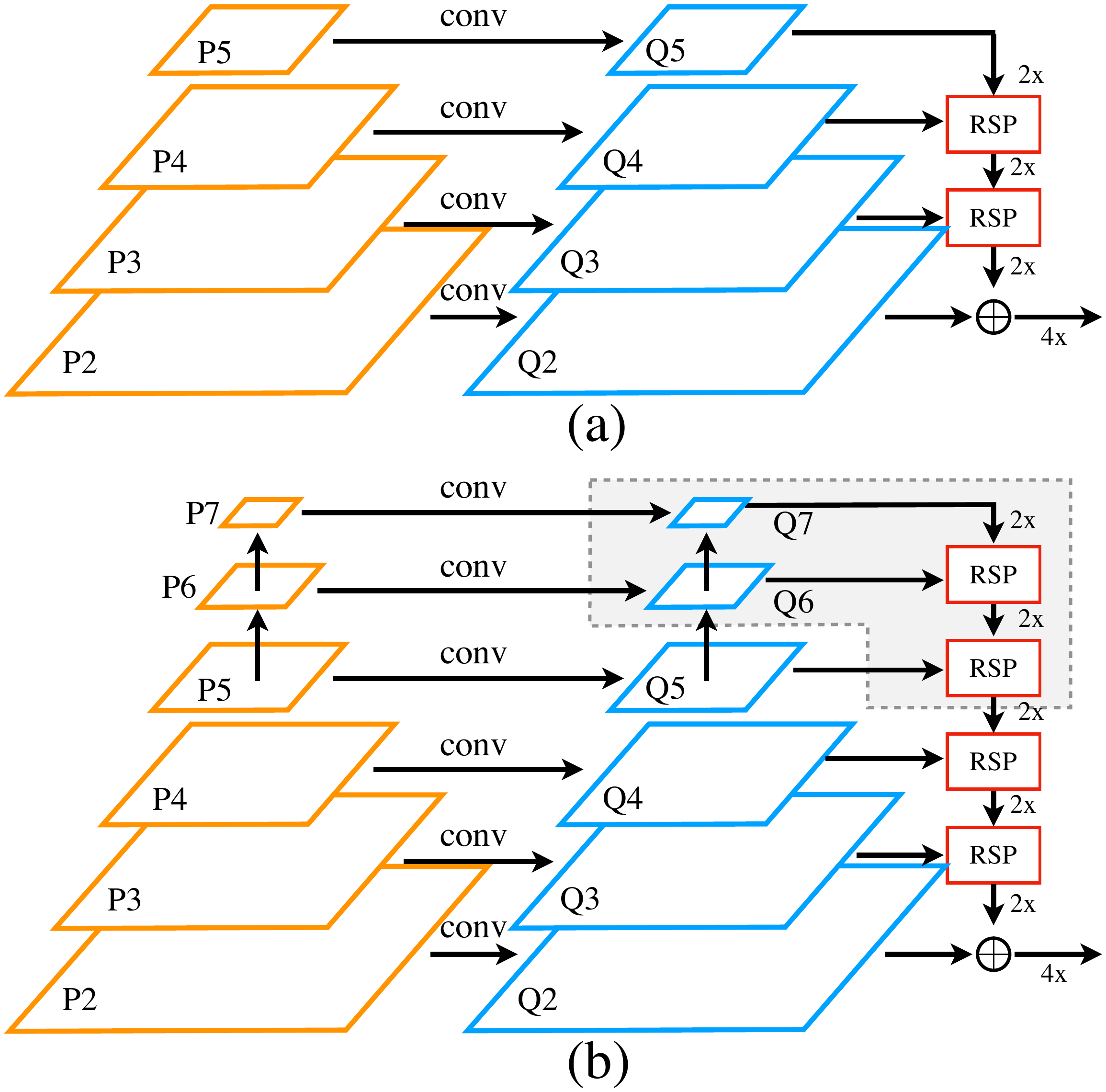}
\end{center}\vspace{-4mm}
  \caption{\textbf{The structure of RSP head.} (a) RSP-2 head. (B) RSP-4 head. To fully exploit the semantic propagation ability of the RSP, we additionally use $\{P_6, P_7\}$ in the FPN. Four RSP modules are connected to aggregate the features from $Q_3$ to $Q_7$, which are transformed from $\{P_3,...,P_7\}$ as in \cite{PanopticFPN}. We do not apply RSP to the fusion of $Q_2$ and $Q_3$ for efficiency. }
\label{fig:rsp_head}
\end{figure}

To this end, we propose a Relational Semantics Extractor (RSE) inspired by \cite{localrelation} to enable the low-level feature to extract the complementary relational context from adjacent high-level feature maps by using a cross-scale pixel-to-region relation operation.
The key insight is that the proposed local relation operation essentially learns the composability between objects in the adjacent feature maps.
The spatial relation is established by adding a positional embedding \cite{relation_modeling}.
On top of RSE, we present the Relational Semantics Propagator (RSP) to propagate the extracted relational context. To progressively propagate the high-level context in a top-down manner, we construct an RSP head by stacking several RSP modules as in Fig. \ref{fig:rsp_head}.
As illustrated in Fig. \ref{fig:story}(c) and (d), our simple and efficient model architecture allows each low-level feature to search and aggregate context information from a large region in the high-level feature map.
The blue arrow and red arrow indicate the context extraction and context propagation respectively, 
together with a top-down progressive contextual information propagation and a large relation operation region. We essentially enable the low-level feature D to capture long-range dependencies from another low-level feature C.
In summary, our contributions are:
\begin{enumerate}
  \item Propose a cross-scale pixel-to-region relation operation as an effective solution to multi-scale feature aggregation.
  \item Propose a Relational Semantics Extractor (RSE) and Relational Semantics Propagator (RSP) for context extraction and propagation respectively.
  \item Introduce a simple, light-weight yet effective RSP head for semantic segmentation which performs competitively on Cityscapes and COCO.
\end{enumerate}

\section{RELATED WORK}
\noindent\textbf{Multi-scale feature aggregation.} Following the earlier work \cite{parsenet}, various successful approaches have been developed based on exploiting multi-scale feature aggregation. Methods like \cite{pspnet, deeplabv2, deeplabv1, deeplabv3} and \cite{deeplabv3+} extract multi-scale features with pyramid pooling and atrous spatial pyramid pooling respectively. On the other hand, Lin et al. \cite{FPN} exploit the natural structure of the deep networks for the construction of multi-scale semantics. A recently proposed network \cite{PanopticFPN} upsamples the multi-scale feature pyramid to the same spatial dimension through lateral paths and fuse them by element-wise summation. Chen et al. \cite{attention2scale} use image pyramids of different scales as input, then use a CNN trunk to fuse multi-scale information by weighted summation. These methods use either concatenation or element-wise summation during fusion which propagates all high-level contextual information to the lower level. Besides, they conduct multi-scale feature aggregation based on a pixel-to-pixel correspondence and do not consider their interrelation. 

\noindent\textbf{Attention.} Attention-based methods have shown great potential in computer vision. Wang et al. \cite{nonlocal} demonstrate that long-range dependencies are beneficial to classification. Parmar et al. \cite{standalone} takes one step further and shows that in the image recognition task, the convolutional kernel can be replaced by a form of self-attention operation. Compositional relationship between pixels in a local neighborhood is exploited in \cite{localrelation, exploring}  to meaningfully join elements together, and it highlights that a meaningful fusion is determined by the similarity of two pixels' feature projections into a learned embedding space \cite{relation_modeling}. Recent work \cite{fpt} applies a non-local operation to compare feature maps from two scale levels for feature enhancement.
Our work extends local relation operation to cross-scale settings to learn the multi-scale composability to achieve a meaningful aggregation of information from multiple scale levels.
A coarse-to-fine approach \cite{acfnet} exploits the coarse prediction to obtain a class center feature as context and then use it to enhance the coarse prediction. Yuan et al. \cite{ocr} aggregates context from object regions in an image through a coarse prediction and distributes it back to all spatial positions based on the relationships between the feature position and the context representation \cite{doubleatten}.
These approaches leverage the pixel-to-region relation to extracting the context but are restrained to a single scale level whereas our method aggregates the related context in a cross-scale setting.
Ding et al. \cite{shape-variant-context} produce a context map for each pixel with a paired convolution and Gaussian kernel in a large predefined region. Then apply the mask to the weights of conv operations to make it shape-variant. Due to the computation cost, the shape-variant context-mask restrained to a single layer of low-resolution.
\begin{figure*}[t]
\begin{center}
  \includegraphics[width=0.9\textwidth]{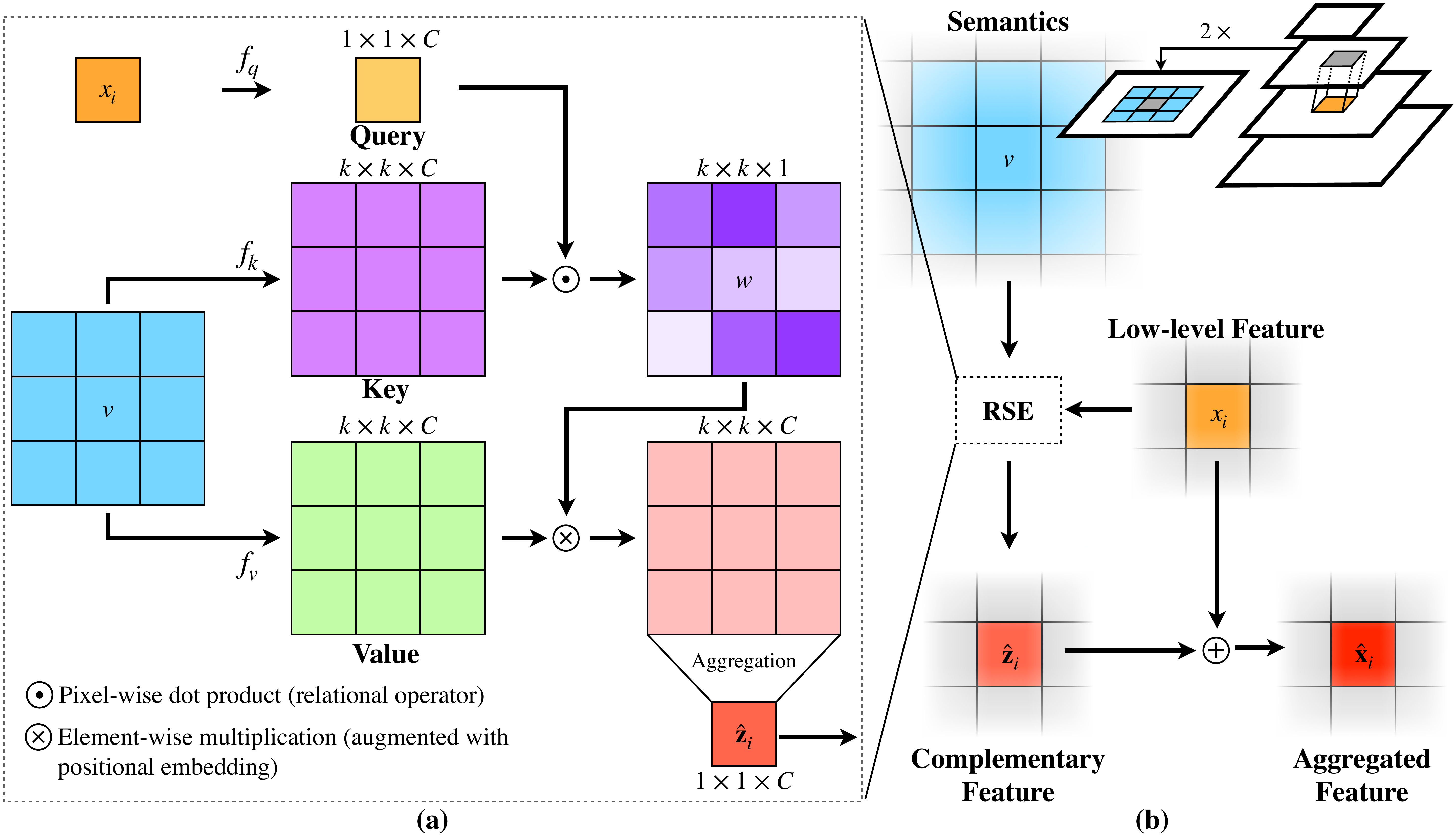}
\end{center}\vspace{-4mm}
  \caption{(a) Cross-scale Relational semantics extractor (RSE) (b) Relational semantics propagator (RSP). The key insight is that the relation operation extracts complementary features from the key and the value and passes them to the query. We exploit this property to extract complementary information from the corresponding region in the high-level feature map w.r.t. to the low-level feature. }
\label{fig:relation_op}
\end{figure*}

\section{APPROACH}
\subsection{Relational Semantics Extractor}
\label{sec:rse}
To address the inefficiency of the convolutional layer in modeling the compositional relationships, \cite{localrelation} propose to explicitly exploit relations between different pixels to extract meaningful features with relation operation. One key insight is that the proposed local relation operation essentially learns the composability between objects in the key map and the query map.
The local relation operation obtains the key and value from the same region. In contrast, we propose a relational semantics extractor (RSE) to exploit the property of relation operation to enable the low-level feature map to selectively extract complementary context from its adjacent high-level feature map with a pixel-to-region correspondence as shown in Fig. \ref{fig:relation_op}. Formally, the operations in relational semantics extractor are defined as follows, given the upsampled high level feature map $\mathbf{z}\in\mathbb{R}^{H\times W\times C}$ and the low level feature map $\mathbf{x}\in\mathbb{R}^{H\times W\times C}$: 
\begin{equation}
    \hat{\mathbf{z}}_i = \Phi(f_q(\mathbf{x}_i), f_k(\mathcal{L}(\mathbf{x}_i, \mathbf{z}))\otimes f_v(\mathcal{L}(\mathbf{x}_i, \mathbf{z}))\
    \label{equ:relational_semantics}
\end{equation}
\noindent where $\hat{\mathbf{z}}\in\mathbb{R}^{H\times W\times C}$ is the output feature map, $\mathbf{x}_i\text{ and }\hat{\mathbf{z}_i}\in\mathbb{R}^{1\times 1\times C}$ is the feature of a specific pixel at location $i$ in $\mathbf{x}$ and $\hat{\mathbf{z}_i}$ respectively. $\Phi$ is the relation operator, which looks for composability between the input pixel $\mathbf{x}_i$ and the defined adjacent region of $\mathbf{x}_i$. $\mathcal{L}(a,\mathbf{b})$ extracts adjacent region of pixel $a$ in feature map $\mathbf{b}$. $f_q, f_k$ and $f_v$ denotes linear transformations that project the features into the embedding space. 
If we define the kernel size of RSE as $k$, $\mathbf{v}=\mathcal{L}(\mathbf{x}_i, \mathbf{z})$ extracts the feature matrix $\mathbf{v}\in\mathbb{R}^{k\times k\times C}$, with the center of $\mathbf{v}$ at the same location with $\mathbf{x}_i$, as visualized in Fig.~\ref{fig:relation_op}(b). $\odot$ is a pixel-wise dot product with broadcasting in channel dimension if required. Following \cite{localrelation,relation_modeling}, we denote $f_q(\mathbf{x}_i), f_k(\mathcal{L}(\mathbf{x}_i, \mathbf{z}))$ and $f_v(\mathcal{L}(\mathbf{x}_i, \mathbf{z}))$ as the \textit{query}, \textit{key} and \textit{value} respectively in Fig.\ref{fig:relation_op}(a). To reduce the computation overhead, we reduce the channel number of the key and query in $f_q$ and $f_k$ by a factor of $d$.
The relation operator $\Phi$ computes appearance composability and is defined as the dot product of the feature pairs: 
\begin{equation}
    \mathbf{w} = \Phi(\mathbf{x}_i, \mathcal{L}(\mathbf{x}_i, \mathbf{z})) = \text{Concat}_{\forall \mathbf{x}_j\in \mathcal{L}(\mathbf{x}_i, \mathbf{z})} (\mathbf{x}_i \cdot \mathbf{x}_j)\
\end{equation}
\noindent where the output of $\Phi$ is a weight matrix $\mathbf{w}\in\mathbb{R}^{k\times k\times 1}$. For simplicity, we omit the linear transformation in the formula. There are other forms of relation modeling but their performance is similar ~\cite{localrelation, exploring}, therefore we adopt the dot product by default for implementation efficiency.

Since the current formulation does not encode positional information and is thus permutation invariant, additional positional embedding is required. We follow a similar strategy for positional embedding in \cite{standalone}. In our case, a normalized 2D relative position map goes through its own linear transformation before the embedding is included in the relation operation. The 2D relative position is generated as $\mathbf{p}\in\mathbb{R}^{k\times k\times C}$, where the first $\frac{C}{2}$ channels are row offset and the second half is column offset. The normalization process projects the values between $-1$ and $1$. The relation operator $\Phi'$ with positional embedding can be now defined as:
\begin{equation}
\small 
    \Phi'(f_q(\mathbf{x}_i), f_k(\mathcal{L}(\mathbf{x}_i, \mathbf{z})) = \Phi(f_q(\mathbf{x}_i), f_k(\mathcal{L}(\mathbf{x}_i, \mathbf{z})) + f_p(\mathbf{p})\
\end{equation}
\noindent where $f_p(\mathbf{p})$ denotes the linear transformation of the relative position map $\mathbf{p}$.

\subsection{Relational Semantics Propagation Head}
\label{sec:rsp}
With the RSE that extracts complementary semantic information to the low-level feature maps, RSP propagates the information to the low-level feature map. With the element-wise addition shown in Fig.~\ref{fig:relation_op}(b), we achieve scale fusion with only selected semantics. Specifically, the aggregation process can be expressed as:

\begin{equation}
    \hat{\mathbf{x}}_i = \mathbf{x}_i + \hat{\mathbf{z}}_i\
\end{equation}
\noindent where the $\hat{\mathbf{z}}_i$ is the extracted relational semantics as in Eq.~\ref{equ:relational_semantics}.

Compared to performing element-wise summation for multi-scale feature aggregation, the proposed RSE has two advantages. (A) During aggregation, summation only considers pixel-wise correspondence, while in RSP, information in a larger semantics region is aggregated to one pixel location from the low-level feature map. (B) Instead of propagating all contextual information from the high level semantic features, RSE selectively extracts useful features with respect to the low-level features. 

We construct the RSP head by stacking a number of RSP modules. The overall structure of the RSP head can be seen in Fig.~\ref{fig:rsp_head}. Since the RSP is able to propagate the high level information to the low level feature maps, we follow \cite{fcos,retinanet} and leverage a $7$-level FPN structure~\cite{FPN}. Specifically, $\{P_2, P_3, P_4, P_5\}$ are generated by connecting a $1\times 1$ convolution to the feature maps of different stages in ResNet~\cite{resnet}, and $\{P6, P7\}$ are obtained by applying strided $3\times 3$ convolution to $P5$ and $P6$ respectively. For more details please refer to \cite{retinanet}.

We denote the transformed feature map from $\{P_2,...P_7\}$ as $\{Q_2, ... Q_7\}$, and progressively aggregate the feature maps from the highest level to the lowest level. The high level feature map is first upscaled by a factor of two before it is fed into the RSP. For clarity, we denote the basic version of the RSP head without $\{Q_6, Q_7\}$ as RSP-2. The element-wise summations between level $Q_5$, $Q_4$ and level $Q_4$, $Q_3$ are replaced by the proposed RSP module. The full RSP head with 4 RSP modules is denoted as RSP-4. All the fusion between two scale levels are replaced with the RSP module except for the one between $Q_3$ and $Q_2$, where we perform only simple summation for avoiding high computations. In our experiments, we also show that the aggregation of higher scale features using RSP yields better results.

\section{EXPERIMENTS}
\subsection{Implementation Details}
\label{sec:baseline_network}

\noindent\textbf{Baseline Network.} The baseline network adopts FPN as the backbone for multi-scale feature extraction.
The baseline for RSP-2 uses $\{Q_2, Q_3, Q_4, Q_5\}$ with strides of $\{4, 8, 16, 32\}$ pixels with respect to the input image. Additional $\{Q_6, Q_7\}$ are used in the baseline for RSP-4 with strides of $\{64, 128\}$ pixels. Our baseline networks aggregate the features with a pixel-to-pixel correspondence. It starts from the highest level $Q_5$(RSP-2)/$Q_7$(RSP-4) and gradually approaches $Q_2$ by upsampling the high level feature map to match the spatial dimension of the following low-level feature map with bilinear upsampling and then apply element-wise summation. A final 1$\times$1 convolution, 4$\times$ bilinear upsampling, and softmax are used to generate the per-pixel class labels at the original image resolution. 

\noindent\textbf{Cityscapes.} The Cityscapes dataset \cite{cityscapes} is tasked for urban scene understanding with 19 categories for semantic segmentation evaluation. The dataset contains 5,000 high resolution pixel-level finely annotated images and 20,000 coarsely annotated images. The finely annotated images are divided into 2,975/500/1,525 images for training, validation and testing.

\noindent\textbf{COCO.} The COCO dataset~\cite{coco} is challenging large scale dataset for computer vision tasks. The panoptic segmentation task \cite{panoptic_semseg} uses all 2017 COCO images with 80 thing and 53 staff classes annotated. As we integrate the proposed semantic segmentation head to the panoptic FPN, we evaluate our approach in the panoptic segmentation task. We use mIoU as the evaluation metric for semantic segmentation and also report PQ, Mask AP and Box AP.

\noindent\textbf{Training details.} 
On Cityscapes, we follow \cite{PanopticFPN} and use SGD with 0.9 momentum with 32 images per mini-batch cropped to a fixed 512$\times$1024 size; the training schedule is 40K/15K/10K updates at learning rates of 0.01/0.0001/0.0001 respectively; a linear learning rate warmup \cite{lr-warmup} over 1000 updates starting from a learning rate of 0.001 is applied; a weight decay of 0.0001 is applied; horizontal flippling, color augmentation \cite{ssd}, and crop bootstrapping \cite{mapillary} are used during training; scale train-time data augmentation resizes an input image from 0.5$\times$ to 2.0$\times$ with a 32 pixel step; BN layers are frozen; no test-time augmentation is used. The evaluation metric is mIoU (mean Intersection-over-Union). On COCO dataset, we use the default Mask R-CNN 1$\times$ training setting \cite{detectron} with scale jitter (shorter image side in [640, 800]).

\noindent\textbf{Loss function.} For semantic segmentation, we use the per-pixel cross entropy loss. For panoptic segmentation, we follow \cite{PanopticFPN} and use the weighted sum of the instance segmentation loss and the semantic segmentation loss, $L = \lambda_i (L_c + L_b + L_m)+\lambda_s L_s$. 
The semantic segmentation loss weight is set to be $\lambda_s=0.5$ and instance segmentation loss weight is set to be $\lambda_i=1$.

\subsection{Performance Comparisons.}
\noindent\textbf{Semantic segmentation. } 
We compare RSP with existing semantic segmentation methods on Cityscapes \emph{val} set.
Only \emph{fine} annotation is used for training and the mIoU is evaluated \emph{without} using flip and multi-scale testing. 
We first compare with Semantic FPN~\cite{panoptic_semseg} on Cityscapes \emph{val} as our RSP head is most similar to the Semantic FPN \cite{PanopticFPN}. The results are shown in Table~\ref{tab:results_in_cityscapes}. 
'D’ in model name indicates use of dilated kernel of size 3 and dilation 3, the detail is in Section \ref{sec:dilated_kernel}.
RSP-2 outperforms Semantic FPN with $\sim$15\% fewer FLOPs. It is worth noting that RSP-4 with ResNet-50-FPN backbone already achieves 77.5\% mIoU, which is very close to the result of Semantic FPN 77.7\% mIoU with the heavier ResNet-101 backbone. 
Next, we compare RSP-4 with other top-performing methods. 
The results are shown in Table \ref{tab:sota_comparison}.
Note that, RSP-4 outperforms DeeplabV3 \cite{deeplabv3} by 0.7\% with 75\% fewer FLOPs when using the same backbone. 
Our approach, which is simple in design, is able to perform on par with DeepLabV3+ which have undergone many design iterations.
RSP-4 achieves strong results compared to state-of-the-art method OCR \cite{ocr} with lighter FLOPs.

\noindent\textbf{Panoptic segmentation.} Next, we conduct experiments to compare with the semantic segmentation branch in the panoptic segmentation task on COCO \emph{val} set by replacing the semantic segmentation branch with the RSP head. The results are shown in Table \ref{tab:panoptic_comparison}. RSP improves the semantic segmentation performance mIoU by a large margin and this also leads to improvement in the panoptic segmentation metric PQ.
\begin{table}[t] \centering
\caption{\bd{Semantic segmentation results on Cityscapes \emph{val} set.} Only \emph{fine} Cityscapes annotations are used for training. 'D' in the model name indicates the use of dilated kernel of size 3 and dilation 3. The median and standard deviation of 5 random runs are reported and the best results are in bold. Note that RSP-4 with ResNet-50-FPN backbone achieves a very close performance to Semantic FPN\cite{PanopticFPN} with ResNet-101-FPN backbone. FLOPs (multiply-adds $\times 10^9$) and the number of parameters are only calculated for the \emph{head} i.e. backbone excluded.\label{tab:results_in_cityscapes}}\vspace{-1mm}
\tablestyle{6pt}{1.0} \footnotesize 
\begin{tabular}{l|c|ccc}
Model                                       & Backbone                    & mIoU                & FLOPs   & \# param.   \\ \shline
\footnotesize Baseline                      & \scriptsize ResNet-50-FPN   & 74.8                & 51.7G   & 4.7M        \\
\footnotesize RSP-2                         & \scriptsize ResNet-50-FPN   & 76.1 $\pm$ 0.2      & 53.4G   & 5.1M        \\
\footnotesize RSP-4                         & \scriptsize ResNet-50-FPN   & \bd{77.5 $\pm$ 0.2} & 53.7G   & 7.8M        \\
\hline
\footnotesize Baseline                      & \scriptsize ResNet-101-FPN  & 76.7                & 51.7G   & 4.7M        \\
\footnotesize RSP-2-D                       & \scriptsize ResNet-101-FPN  & 77.9 $\pm$ 0.2      & 53.4G   & 5.1M        \\
\footnotesize RSP-4-D                       & \scriptsize ResNet-101-FPN  & \bd{78.5 $\pm$ 0.2} & 53.7G   & 7.8M        \\
\shline
\scriptsize Semantic FPN\cite{PanopticFPN}  & \scriptsize ResNet-50-FPN   & 75.8                & 62.5G   & 6.5M        \\
\scriptsize Semantic FPN\cite{PanopticFPN}  & \scriptsize ResNet-101-FPN  & 77.7                & 62.5G   & 6.5M        \\
\end{tabular}
\end{table}

\begin{table}[b] \centering
\tablestyle{5pt}{1.0} 
\caption{\bd{Performance comparisons on Cityscapes \emph{val} set.} Only \emph{fine} Cityscapes annotations are used for training. The mIoU is evaluated \emph{w/o} using ﬂip and multi-scale testing. 'D' in model name indicates use of dilated kernel of size 3 and dilation 3. The backbone notation includes the dilated resolution 'D'. FLOPs (multiply-adds) and memory (\# activations) are calculated for the \emph{whole} model i.e. includes backbone and head. Memory are approximate but informative.}
\label{tab:sota_comparison}\vspace{-1mm}
\begin{tabular}{l|c|lcc}
Model                                       & Backbone                      & mIoU      & FLOPs      & memory.      \\ \shline
\footnotesize Semantic FPN\cite{PanopticFPN}& \scriptsize ResNet-101-FPN    & 77.7      & 0.5T       & 0.8G         \\
\footnotesize DeeplabV3 \cite{deeplabv3}    & \scriptsize ResNet-101-D8     & 77.8      & 1.9T       & 1.9G         \\
\footnotesize PSANet101 \cite{psanet}       & \scriptsize ResNet-101-D8     & 77.9      & 2.0T       & 2.0G         \\
\footnotesize SETR-PUP  \cite{setr}         & \scriptsize T-Large           & 79.3      & 1.0T      & 2.7G          \\
\footnotesize Mapillary \cite{mapillary}    & \scriptsize WideResNet-38-D8  & 79.4      & 4.3T       & 1.7G         \\
\footnotesize DeeplabV3+ \cite{deeplabv3+}  & \scriptsize X-71-D16          & 79.6      & 0.5T       & 1.9G         \\
\footnotesize OCR       \cite{ocr}          & \scriptsize HRNetV2           & 80.8      & 1.3T       & 1.4G         \\
\hline
\footnotesize \bd{RSP-4-D}                  & \scriptsize ResNet-101-FPN    & 78.5      & 0.5T       & 0.8G         \\
\footnotesize \bd{RSP-4-D}                  & \scriptsize ResNeXt-101-FPN   & 79.5      & 0.8T       & 1.4G         \\
\end{tabular}
\end{table}

\begin{table}[t] \centering
\tablestyle{3.5pt}{1.0} \footnotesize
\caption{\bd{Panoptic segmentation results on COCO \emph{val}.} The backbone is ResNet-50-FPN. 'D' in model name indicates use of dilated kernel of size 3 and dilation 3. The backbone notation includes the dilated resolution 'D'. 
In the second row and third row we replace the original semantic segmentation branch in the Panoptic FPN with our RSP-2-D and RSP-4-D head respectively. The FLOPs (multiply-adds $\times 10^9$) and number of parameters are calculated for the head.  only.\label{tab:panoptic_comparison}}\vspace{-1mm}
\begin{tabular}{l|cccc|cc}
Model                                           & mIoU      & PQ        & Mask AP & Box AP  &FLOPs  & \# param\\
\shline
\footnotesize Panoptic FPN \cite{PanopticFPN}   & 41.3      & 39.4      & 34.6    & 37.5    & 62.5G & 6.5M \\
\footnotesize RSP-2-D head                      & 41.9      & 40.1      & 34.6    & 37.5    & 53.1G & 5.1M \\
\footnotesize RSP-4-D head                      & \bd{42.7} & \bd{40.2} & 34.5    & 37.5    & 53.4G & 7.8M \\
\end{tabular}
\end{table}

\begin{table}[b]\centering
\tablestyle{4.5pt}{1.2} \footnotesize
\caption{\textbf{Effect of the number of RSP modules in the RSP-2 head with backbone ResNet-50-FPN.}  $\times$1 and $\times$2 indicates the number of RSP modules employed. In column RSP and Sum, (54,43) means employing RSP/Sum between level $Q_5$, $Q_4$ and $Q_4$, $Q_3$.
In \textit{+ SELF$\times 2$}, we replace the cross-scale relation operation with the local relation operation \cite{localrelation}. In \textit{+ CONTEXT$\times 2$} we replace the pair-wise relation operation in RSE with the averaging operation. 
$^*$For reference purpose. We re-train the semantic FPN~\cite{PanopticFPN} with the same training settings as our RSP head.}
\label{tab:rsp_head}\vspace{-1mm}
\begin{tabular}{l|cc|ccc}
Model                                    & RSP     & Sum       & mIoU                  & FLOPs & \# param.\\ \shline
 BASELINE                                & -        & (54, 43)  & 74.8                  & 51.7G & 4.7M  \\
+ RSP$\times 1$                          & 54       & 43        & 75.2\dt{+0.4}         & 52.0G & 4.9M  \\
+ RSP$\times 1$                          & 43       & 54        & 75.6\dt{+0.8}         & 53.1G & 4.9M  \\
+ RSP$\times 2$                          & (54, 43) & -         & \bd{76.1\dt{+1.3}}    & 53.4G & 5.1M \\
+ SELF$\times 2$                         & -        & -         & 75.5\dt{+0.7}         & 53.4G & 5.1M \\ 
+ CONTEXT$\times 2$                      & - & - & 75.2\dt{+0.4} & 51.7G & 4.7M
\end{tabular}
\end{table}
\begin{table}[t]\centering
\caption{\bd{The effect of the kernel sizes and dimension reduction factor $d$ in the RSP module.} Experiments are conducted with RSP-2 head on the backbone ResNet-50-FPN.  \label{tab:rsp_module_ablation}}\vspace{-1mm}
\subfloat[\bd{Performance with different RSE kernel size.} We alter the kernel size and the dilation of the kernels in the RSE to discover the optimal setting. K and D indicates the kernel size and dilation respectively.]
{\tablestyle{15pt}{1.0}\begin{tabular}{c|ccc}
(K, D)      & mIoU          & FLOPs             & \# param. \\ 
\shline
(3, 1)      & 75.5          & 53.1G             & 5.1M  \\
(5, 1)      & 75.5          & 53.2G             & 5.1M  \\
(7, 1)      & \textbf{76.1} & 53.4G             & 5.1M  \\
(3, 2)      & 75.6          & 53.1G             & 5.1M  \\
(3, 3)      & 76.0          & 53.1G             & 5.1M  \\
\end{tabular}}\\
\subfloat[\bd{Performance with different number of middle channels.} The output channels of the value transformation $f_v$ are 128. As mentioned before, dimension reduction is applied in $f_q$ and $f_v$ to reduce the channel number by a factor of $d$ before further operations. We indicate the number of middle channels in brackets.]
{\tablestyle{15pt}{1.0}\begin{tabular}{c|ccc}
$d$ & mIoU      & FLOPs         & \# param.                         \\ \shline
1 (128)         & 75.9          & 53.8G         & 5.3M             \\
2 (64)          & \textbf{76.1} & 53.4G         & 5.1M             \\
4 (32)          & 75.6          & 53.0G         & 5.0M             \\
8 (16)          & 76.0          & 52.8G         & 4.9M                            
\end{tabular}}\hspace{2mm}
\end{table}
\begin{table}[b]\centering
\caption{\textbf{Effect of the number of RSP module in the RSP-4 head with backbone ResNet-50-FPN.} $\times$1-4 indicates the number of RSP module employed. In column RSP and Sum, (54,43) means employing RSP/Sum between level $Q_5$, $Q_4$ and $Q_4$, $Q_3$. \label{tab:rsp-4}}
\vspace{-1mm}
\tablestyle{1.2pt}{1.2} \footnotesize
\begin{tabular}{l|cc|ccc}
Model                   & RSP           & Sum               & mIoU              & FLOPs     & \# param. \\ \shline
\scriptsize BASELINE    & -             & (54, 43)          & 74.8              & 51.7G     & 4.7M      \\
+ Q6, Q7                & -             & (76, 65, 54, 43)  & 76.0\dt{+1.2}     & 51.9G     & 7.1M      \\
\hline
+ RSP$\times 1$         & 43            & (76, 65, 54)      & 76.3\dt{+1.5}     & 53.2G     & 7.3M  \\
+ RSP$\times 1$         & 76            & (65, 54, 43)      & 76.2\dt{+1.4}     & 51.9G     & 7.3M  \\
\hline
+ RSP$\times 2$         & (54, 43)      & (76, 65)          & 76.4\dt{+1.6}     & 53.6G     & 7.4M     \\
+ RSP$\times 2$         & (76, 65)      & (54, 43)          & 76.9\dt{+2.1}     & 52.0G     & 7.4M \\
\hline
+ RSP$\times 3$         & (65, 54, 43)  & 76                & 77.1\dt{+2.3}     & 53.7G     & 7.6M      \\
+ RSP$\times 3$         & (76, 65, 54)  & 43                & 77.2\dt{+2.4}     & 52.3G     & 7.6M  \\
\hline
+ RSP$\times 4$         & (76, 65, 54, 43) & -              & \bd{77.5\dt{+2.7}} & 53.7G & 7.8M
\end{tabular}
\end{table}

\subsection{Ablation Study on Cityscapes}
\label{sec:dilated_kernel}
\noindent\textbf{Ablation study of the RSP-2 head.} We break down the improvements of RSP-2 over the baseline, by adding RSP modules to the baseline one-by-one. The results are shown in Table \ref{tab:rsp_head}. Adding RSP module consistently improves the baseline. With 2 RSP modules ($\sim$3\% computation increment), the RSP head achieves a 1.3 mIoU improvement over the baseline. 
In the experiment \textit{+ SELF}, we replace all the cross-scale relation operations with local relation operation \cite{localrelation}, and the input is only the high-level feature map. 
Compared to \textit{+ SELF}, RSP achieves much better performance because the cross-scale setting of our relation operation enables the low-level feature to access context from a much larger region.
In the experiment \textit{+ CONTEXT}, we propagate high-level semantic information by simply aggregating high-level features in a local receptive field by average pooling and add it to the low-level feature. This outperforms the baseline but not our RSP-2. It proves the superiority of our proposed relation operation in extracting meaningful context information from the high-level feature map.

\noindent\textbf{Ablation study of RSP Module.} We analyze the effect of kernel sizes in the RSP module, as shown in Table~\ref{tab:rsp_module_ablation}. The RSP-2 achieves the best performance when the kernel size is 7 and dilation is 1. Meanwhile, RSP obtains a similar result for kernel size 3 and dilation 3, where the effective kernel size is also 7. Therefore, we decide to adopt kernel size 7 and dilation 1 when using backbone ResNet-50 for a better performance, and kernel size 3 and dilation 3 when using larger backbone ResNet-101 since using the dilation reduces the number of computation as well as the GPU memory.  For the number of middle channels, we choose the dimension reduction factor as $2$ for its better performance. 

\noindent\textbf{Ablation study of RSP-4 head.} We study effect of the number of RSP modules in the RSP-4 head, results are in Table~\ref{tab:rsp-4}.
We have three observations. 1) Increasing the depth improves the performance even the additional higher-level features are aggregated by element-wise summation. This confirms that high-level semantics is beneficial to classification. 2) Increasing the number of RSP modules consistently improves the performance. 3) With the same number of RSP modules employed, start adding RSP modules from the highest-level generally gives a better result than from the lowest level.
This proves that the proposed RSP successfully meets our design goal to propagates the complementary contextual information in a top-down manner.
\subsection{Qualitative Evaluation.} 
\begin{figure}[t]
\begin{center}
  \includegraphics[width=1\linewidth]{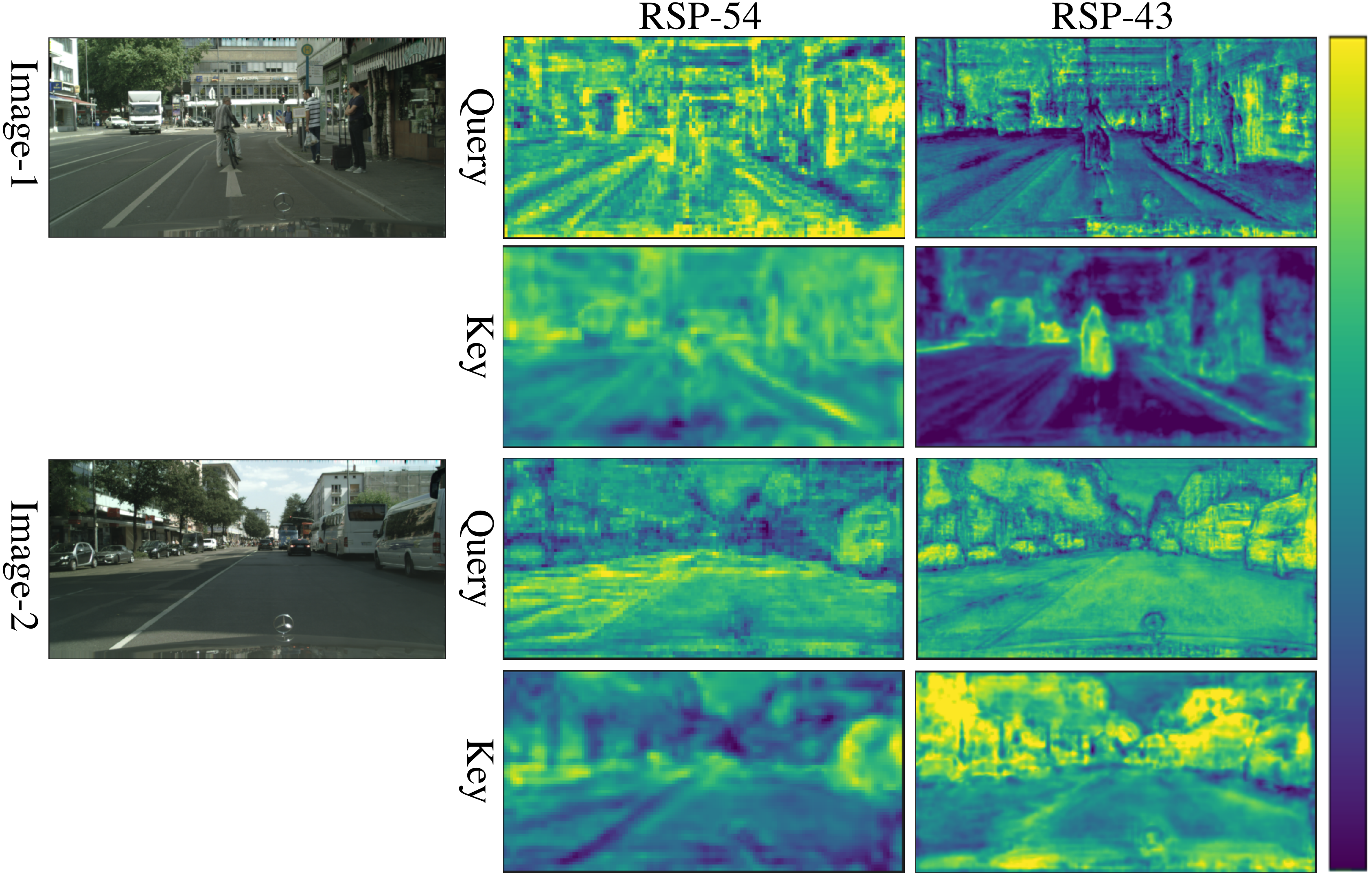}
\end{center}\vspace{-4mm}
  \caption{\textbf{Illustration of learnt \textit{key} and \textit{query}.} RSP-AB indicates RSP module placed between $Q_A$ and $Q_B$. The complementary property between the query and key features visualised here is the core insight that we leverage to extract complementary information \textit{w.r.t} the query map from the value map.}
\label{fig:visualise_query_key}
\end{figure}

\noindent\textbf{Complementary information in relation operation.} As shown in \cite{localrelation}, the feature extracted by the query and key transformation complement each other. In our case, we demonstrate that in cross-scale relation operation, this observation stands. As is visualized in Fig.~\ref{fig:visualise_query_key}, both key-query feature pairs in RSP-54 and RSP-43 complement each other.

\begin{figure*}[t]
\begin{center}
  \includegraphics[width=0.95\textwidth]{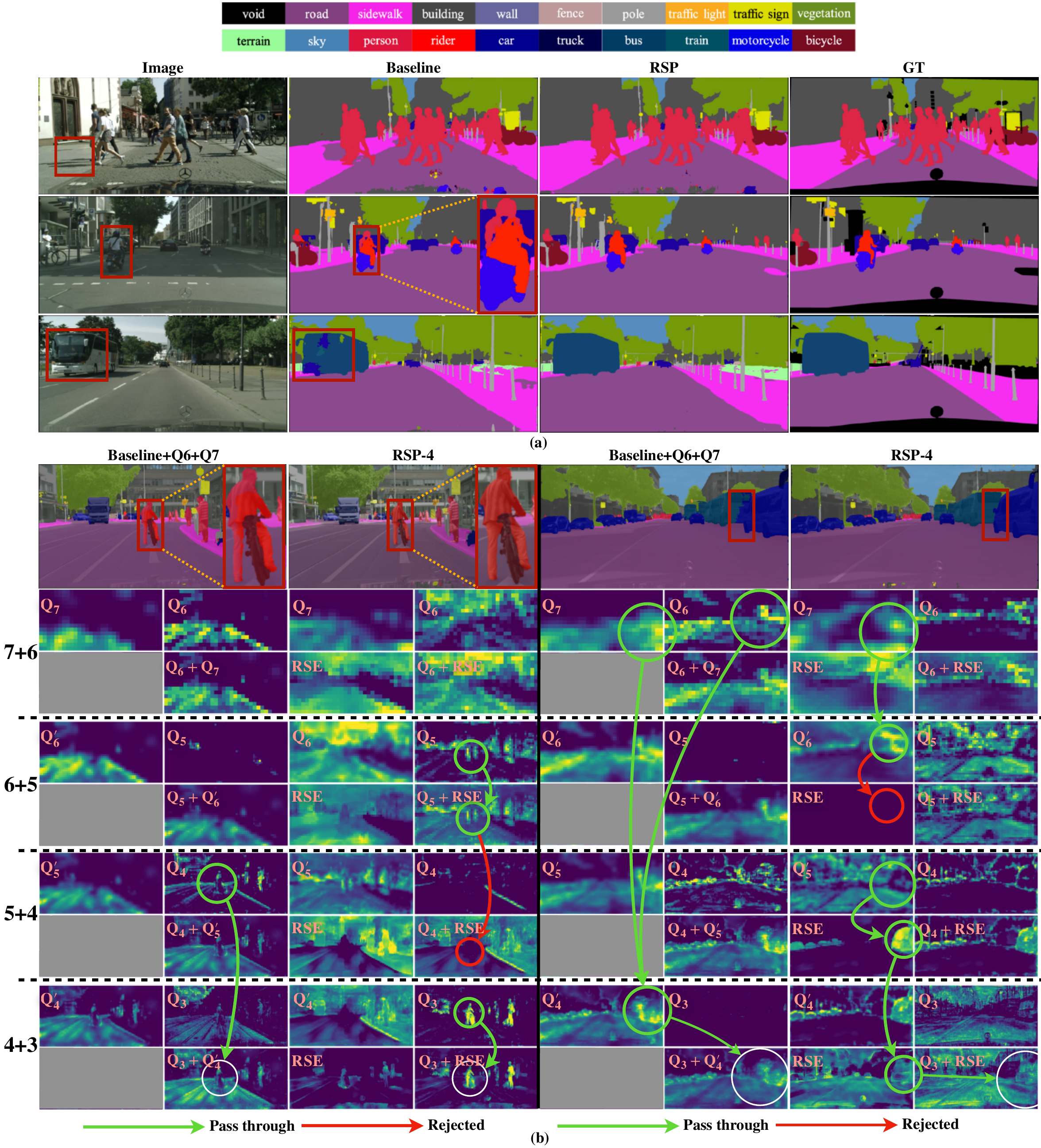}
\end{center}\vspace{-4mm}
  \caption{
  (a)\textbf{ Qualitative results on Cityscapes.}
  The challenging area for the baseline model is at the places where a transition from one class to another class occurs consequently multiple context information is available in those areas. The ability to select the right context in our proposed RSP-4 plays a key role in making the correct classification. 
  (b)\textbf{ Visualization of the context propagation.}
  We visualize and compare the feature maps from the same channel produced by RSP-4 and the baseline during the whole feature aggregation process on two images. $Q_x$ follows the notation defined before \ref{sec:baseline_network}. RSE box shows the features extracted by the RSE operation. Gray box means no relational features are extracted. 
  $Q_{x}+Q_{x+1}\text{/RSE}$ box displays the aggregated cross-scale features.
  $Q'_x$ box shows the upsampled aggregated features, which means $Q'_x=\text{upsampling}(Q_{x}+Q_{x+1}\text{/RSE})$. 
  The fusion between $Q_3$ and $Q_2$ is not visualized as the structures of RSP and the baseline are the same in this part. The green path indicates the passage of features, while the red path indicates the areas where the high-level semantic features are rejected by RSE.}
\label{fig:qualitative_comparison}
\end{figure*}

\noindent\textbf{Qualitative results on Cityscapes.}
We provide the qualitative comparisons between the RSP-4 and the baseline network with ResNet-50-FPN in the upper part of Fig.~\ref{fig:qualitative_comparison}(a). We use the \emph{red} box to mark those challenging regions. 
The baseline model misclassifies the sidewalk near the crowd as the road in the first image, the portion of the rider far from the motorcycle as a person in the second image, and pixels at the boundary of a bus as the car in the last image. In contrast, the proposed RSP head classifies all those areas correctly. The rider case demonstrates that the RSP head enables long-range dependencies to be captured. The sidewalk and bus case confirms that the RSP head allows the pixel at the boundary to select the helpful high-level context.

\noindent\textbf{Visualization of the context propagation.}
We visualize and compare the feature maps from the same channel produced by RSP-4 and the baseline during the whole feature aggregation process on two images Fig.~\ref{fig:qualitative_comparison}(b). In the left image, half part of the rider not near the bicycle is misclassified as the person. In the right image, the pixels at the boundary of the car and bus are misclassified. 
The feature maps that display the context propagation show that the baseline model fully passes down the high-level context which includes wrong or incomplete context whereas the RSP-4 successfully reject those context and aggregate the complementary and informative context.
In the last row, we use white circles to highlight the features produced by the RSP-4 and baseline model that represents rider and car in the red box. The RSP-4 produces complete and clear features which is much easier to be discriminated against.

\section{CONCLUSION}
In this work, we propose a relational multi-scale feature aggregation approach for semantic segmentation. 
The multi-scale feature aggregation is achieved through the proposed relational semantics propagator (RSP) head, where the high-level context is selectively propagated to the low-level feature maps with a pixel-to-region correspondence. 
We propose a cross-scale relation operation named relational semantics extractor (RSE) to extract complementary contextual information w.r.t. the low-level feature from the corresponding region of adjacent high-level feature maps.
The cross-scale setting also enables the low-level features to capture long-range dependency in a compute-efficient way.
Extensive experiments show the effectiveness of the RSP module and the consistent improvement by adding multiple RSP modules. 


\section*{ACKNOWLEDGMENT}

We thank Feng Xue and Guirong Zhuo for their helpful discussion and generous support and the Institute of Intelligent Vehicles, School of Automotive Studies, Tongji University for providing the GPUs for experiments.

\bibliographystyle{./IEEEtran} 
\bibliography{./IEEEabrv,./relational_scale_aggregation}

\begin{thebibliography}{10}
\providecommand{\url}[1]{#1}
\csname url@rmstyle\endcsname
\providecommand{\newblock}{\relax}
\providecommand{\bibinfo}[2]{#2}
\providecommand\BIBentrySTDinterwordspacing{\spaceskip=0pt\relax}
\providecommand\BIBentryALTinterwordstretchfactor{4}
\providecommand\BIBentryALTinterwordspacing{\spaceskip=\fontdimen2\font plus
\BIBentryALTinterwordstretchfactor\fontdimen3\font minus
  \fontdimen4\font\relax}
\providecommand\BIBforeignlanguage[2]{{%
\expandafter\ifx\csname l@#1\endcsname\relax
\typeout{** WARNING: IEEEtran.bst: No hyphenation pattern has been}%
\typeout{** loaded for the language `#1'. Using the pattern for}%
\typeout{** the default language instead.}%
\else
\language=\csname l@#1\endcsname
\fi
#2}}

\bibitem{deeplabv3}
L.~Chen, G.~Papandreou, F.~Schroff, and H.~Adam, ``Rethinking atrous
  convolution for semantic image segmentation,'' \emph{arXiv preprint
  arXiv:1706.05587}, 2017.

\bibitem{parsenet}
W.~Liu, A.~Rabinovich, and A.~Berg, ``Parse{N}et: Looking wider to see
  better,'' \emph{arXiv preprint arXiv:1506.04579}, 2015.

\bibitem{pspnet}
H.~Zhao, J.~Shi, X.~Qi, X.~Wang, and J.~Jia, ``Pyramid scene parsing network,''
  in \emph{Proceedings of the IEEE conference on computer vision and pattern
  recognition}, 2017, pp. 2881--2890.

\bibitem{deeplabv3+}
L.~Chen, Y.~Zhu, G.~Papandreou, F.~Schroff, and H.~Adam, ``Encoder-decoder with
  atrous separable convolution for semantic image segmentation,'' in
  \emph{Proceedings of the European conference on computer vision (ECCV)},
  2018, pp. 801--818.

\bibitem{acfnet}
F.~Zhang, Y.~Chen, Z.~Li, Z.~Hong, J.~Liu, F.~Ma, J.~Han, and E.~Ding,
  ``{ACFN}et: Attentional class feature network for semantic segmentation,'' in
  \emph{Proceedings of the IEEE International Conference on Computer Vision},
  2019, pp. 6798--6807.

\bibitem{ocr}
Y.~Yuan, X.~Chen, and J.~Wang, ``Object-contextual representations for semantic
  segmentation,'' \emph{arXiv preprint arXiv:1909.11065}, 2019.

\bibitem{hierarchy}
A.~Tao, K.~Sapra, and B.~Catanzaro, ``Hierarchical multi-scale attention for
  semantic segmentation,'' \emph{arXiv preprint arXiv:2005.10821}, 2020.

\bibitem{shape-variant-context}
H.~Ding, X.~Jiang, B.~Shuai, A.~Q. Liu, and G.~Wang, ``Semantic correlation
  promoted shape-variant context for segmentation,'' in \emph{Proceedings of
  the IEEE/CVF Conference on Computer Vision and Pattern Recognition}, 2019,
  pp. 8885--8894.

\bibitem{ccnet}
Z.~Huang, X.~Wang, L.~Huang, C.~Huang, Y.~Wei, and W.~Liu, ``{CCN}et:
  Criss-cross attention for semantic segmentation,'' in \emph{Proceedings of
  the IEEE/CVF International Conference on Computer Vision (ICCV)}, October
  2019.

\bibitem{localrelation}
H.~Hu, Z.~Zhang, Z.~Xie, and S.~Lin, ``Local relation networks for image
  recognition,'' in \emph{Proceedings of the IEEE International Conference on
  Computer Vision}, 2019, pp. 3464--3473.

\bibitem{relation_modeling}
P.~W. Battaglia, J.~B. Hamrick, V.~Bapst, A.~Sanchez-Gonzalez, V.~Zambaldi,
  M.~Malinowski, A.~Tacchetti, D.~Raposo, A.~Santoro, R.~Faulkner,
  \emph{et~al.}, ``Relational inductive biases, deep learning, and graph
  networks,'' \emph{arXiv preprint arXiv:1806.01261}, 2018.

\bibitem{PanopticFPN}
A.~Kirillov, R.~Girshick, K.~He, and P.~Doll{\'a}r, ``Panoptic feature pyramid
  networks,'' in \emph{Proceedings of the IEEE Conference on Computer Vision
  and Pattern Recognition}, 2019, pp. 6399--6408.

\bibitem{deeplabv2}
L.~Chen, G.~Papandreou, I.~Kokkinos, K.~Murphy, and A.~L. Yuille, ``Deep{L}ab:
  Semantic image segmentation with deep convolutional nets, atrous convolution,
  and fully connected {CRF}s,'' \emph{IEEE transactions on pattern analysis and
  machine intelligence}, vol.~40, no.~4, pp. 834--848, 2017.

\bibitem{deeplabv1}
L.~Chen, G.~Papandreou, I.~Kokkinos, K.~Murphy, and A.~L. Yuille, ``Semantic
  image segmentation with deep convolutional nets and fully connected {CRF}s,''
  \emph{arXiv preprint arXiv:1412.7062}, 2014.

\bibitem{FPN}
T.-Y. Lin, P.~Doll{\'a}r, R.~Girshick, K.~He, B.~Hariharan, and S.~Belongie,
  ``Feature pyramid networks for object detection,'' in \emph{Proceedings of
  the IEEE conference on computer vision and pattern recognition}, 2017, pp.
  2117--2125.

\bibitem{attention2scale}
L.~Chen, Y.~Yang, J.~Wang, W.~Xu, and A.~L. Yuille, ``Attention to {S}cale:
  Scale-aware semantic image segmentation,'' in \emph{Proceedings of the IEEE
  conference on computer vision and pattern recognition}, 2016, pp. 3640--3649.

\bibitem{nonlocal}
X.~Wang, R.~Girshick, A.~Gupta, and K.~He, ``Non-local neural networks,'' in
  \emph{Proceedings of the IEEE conference on computer vision and pattern
  recognition}, 2018, pp. 7794--7803.

\bibitem{standalone}
N.~Parmar, P.~Ramachandran, A.~Vaswani, I.~Bello, A.~Levskaya, and J.~Shlens,
  ``Stand-alone self-attention in vision models,'' in \emph{Advances in Neural
  Information Processing Systems}, 2019, pp. 68--80.

\bibitem{exploring}
H.~Zhao, J.~Jia, and V.~Koltun, ``Exploring self-attention for image
  recognition,'' in \emph{Proceedings of the IEEE/CVF Conference on Computer
  Vision and Pattern Recognition}, 2020, pp. 10\,076--10\,085.

\bibitem{fpt}
D.~Zhang, H.~Zhang, J.~Tang, M.~Wang, X.~Hua, and Q.~Sun, ``Feature pyramid
  transformer,'' \emph{arXiv preprint arXiv:2007.09451}, 2020.

\bibitem{doubleatten}
Y.~Chen, Y.~Kalantidis, J.~Li, S.~Yan, and J.~Feng, ``${A}^2$-{N}ets: Double
  attention networks,'' in \emph{Advances in neural information processing
  systems}, 2018, pp. 352--361.

\bibitem{fcos}
Z.~Tian, C.~Shen, H.~Chen, and T.~He, ``{FCOS}: Fully convolutional one-stage
  object detection,'' in \emph{Proceedings of the IEEE international conference
  on computer vision}, 2019, pp. 9627--9636.

\bibitem{retinanet}
T.-Y. Lin, P.~Goyal, R.~Girshick, K.~He, and P.~Doll{\'a}r, ``Focal loss for
  dense object detection,'' in \emph{Proceedings of the IEEE international
  conference on computer vision}, 2017, pp. 2980--2988.

\bibitem{resnet}
K.~He, X.~Zhang, S.~Ren, and J.~Sun, ``Deep residual learning for image
  recognition,'' in \emph{Proceedings of the IEEE conference on computer vision
  and pattern recognition}, 2016, pp. 770--778.

\bibitem{cityscapes}
M.~Cordts, M.~Omran, S.~Ramos, T.~Rehfeld, M.~Enzweiler, R.~Benenson,
  U.~Franke, S.~Roth, and B.~Schiele, ``The cityscapes dataset for semantic
  urban scene understanding,'' in \emph{Proceedings of the IEEE conference on
  computer vision and pattern recognition}, 2016, pp. 3213--3223.

\bibitem{coco}
T.-Y. Lin, M.~Maire, S.~Belongie, J.~Hays, P.~Perona, D.~Ramanan,
  P.~Doll{\'a}r, and C.~L. Zitnick, ``Microsoft {COCO}: Common objects in
  context,'' in \emph{European conference on computer vision}.\hskip 1em plus
  0.5em minus 0.4em\relax Springer, 2014, pp. 740--755.

\bibitem{panoptic_semseg}
A.~Kirillov, K.~He, R.~Girshick, C.~Rother, and P.~Doll{\'a}r, ``Panoptic
  segmentation,'' in \emph{Proceedings of the IEEE conference on computer
  vision and pattern recognition}, 2019, pp. 9404--9413.

\bibitem{lr-warmup}
P.~Goyal, P.~Doll{\'a}r, R.~Girshick, P.~Noordhuis, L.~Wesolowski, A.~Kyrola,
  A.~Tulloch, Y.~Jia, and K.~He, ``Accurate, {L}arge {M}inibatch {SGD}:
  Training imagenet in 1 hour,'' \emph{arXiv preprint arXiv:1706.02677}, 2017.

\bibitem{ssd}
W.~Liu, D.~Anguelov, D.~Erhan, C.~Szegedy, S.~Reed, C.-Y. Fu, and A.~C. Berg,
  ``{SSD: Single shot multibox detector},'' in \emph{European conference on
  computer vision}.\hskip 1em plus 0.5em minus 0.4em\relax Springer, 2016, pp.
  21--37.

\bibitem{mapillary}
S.~Rota~Bul{\`o}, L.~Porzi, and P.~Kontschieder, ``In-place activated batchnorm
  for memory-optimized training of {DNN}s,'' in \emph{Proceedings of the IEEE
  Conference on Computer Vision and Pattern Recognition}, 2018, pp. 5639--5647.

\bibitem{detectron}
R.~Girshick, I.~Radosavovic, G.~Gkioxari, P.~Doll\'{a}r, and K.~He,
  ``Detectron,'' \url{https://github.com/facebookresearch/detectron}, 2018.

\bibitem{psanet}
H.~Zhao, Y.~Zhang, S.~Liu, J.~Shi, C.~Change~Loy, D.~Lin, and J.~Jia,
  ``{PSANet: Point-wise spatial attention network for scene parsing},'' in
  \emph{Proceedings of the European Conference on Computer Vision (ECCV)},
  2018, pp. 267--283.

\bibitem{setr}
S.~Zheng, J.~Lu, H.~Zhao, X.~Zhu, Z.~Luo, Y.~Wang, Y.~Fu, J.~Feng, T.~Xiang,
  P.~H. Torr, \emph{et~al.}, ``{R}ethinking semantic segmentation from a
  sequence-to-sequence perspective with transformers,'' \emph{arXiv preprint
  arXiv:2012.15840}, 2020.

\end{thebibliography}

\end{document}